\documentclass{article}

\usepackage{arxiv}

\usepackage[utf8]{inputenc} % allow utf-8 input
\usepackage[T1]{fontenc}    % use 8-bit T1 fonts
\usepackage{hyperref}       % hyperlinks
\usepackage{url}            % simple URL typesetting
\usepackage{booktabs}       % professional-quality tables
\usepackage{amsfonts}       % blackboard math symbols
\usepackage{nicefrac}       % compact symbols for 1/2, etc.
\usepackage{microtype}      % microtypography
\usepackage{lipsum}		% Can be removed after putting your text content
\usepackage{graphicx}
\usepackage{natbib}
\usepackage{doi}

\usepackage{algorithm}
\usepackage{pifont}
\newcommand{\cmark}{\ding{51}}
\newcommand{\xmark}{\ding{55}}

\usepackage{enumerate}
\usepackage[table]{xcolor} 
\usepackage{amsmath,amssymb}
\usepackage{algpseudocode}

\DeclareMathOperator{\Grad}{\partial\!}  % 偏导
\newcommand{\Model}{\mathcal{M}}
\newcommand{\Loss}{\mathcal{L}}
\newcommand{\Gen}{\mathcal{G}}

\definecolor{color_blue}{HTML}{E7EFFA}
\definecolor{color_green}{HTML}{E6F8E0}
\definecolor{color_gray}{HTML}{ECECEC}
\definecolor{pearDark}{HTML}{2980B9}
\usepackage{multirow}

\title{EmoFeedback\textsuperscript{2}: Reinforcement of Continuous Emotional Image Generation via LVLM-based Reward and Textual Feedback}

\date{} 					% Or removing it

\author{
Kai Shu \textsuperscript{1} \thanks{Equal contribution.},\hspace{1em}
Jingyang Jia \textsuperscript{1} \footnotemark[1],\hspace{1em}
Gang Yang\textsuperscript{1},\hspace{1em}
Long Xing\textsuperscript{1,2},\hspace{1em}
Xun Chen\textsuperscript{1},\hspace{1em}
Aiping Liu\textsuperscript{1} \thanks{Corresponding author.} \\
\textsuperscript{1} University of Science and Technology of China \\
\textsuperscript{2} Shanghai Artificial Intelligence Laboratory\\
{\tt\small \{sk18872125513@, jjygood@\}mail.ustc.edu.cn} \\
}

\renewcommand{\shorttitle}{\textit{arXiv} Template}

\hypersetup{
pdftitle={A template for the arxiv style},
pdfsubject={q-bio.NC, q-bio.QM},
pdfauthor={David S.~Hippocampus, Elias D.~Striatum},
pdfkeywords={First keyword, Second keyword, More},
}

\begin{document}
\maketitle

\begin{abstract}
Continuous emotional image content generation (C-EICG) is emerging rapidly due to its ability to produce images aligned with both user descriptions and continuous emotional values. However, existing approaches lack emotional feedback from generated images, limiting the control of emotional continuity. 
Additionally, their simple emotion-text alignment fails to adaptively adjust emotional prompts according to image content, leading to insufficient emotional fidelity. To address these concerns, we propose a novel generation-understanding-feedback reinforcement paradigm (\textbf{EmoFeedback\textsuperscript{2}}) for C-EICG, which exploits the reasoning capability of the fine-tuned large vision–language model (LVLM) to provide reward and textual feedback for generating high-quality images with continuous emotions. 
Specifically, we introduce an \textit{emotion-aware reward feedback} strategy, where the LVLM evaluates the emotional values of generated images and computes the reward against target emotions, guiding the reinforcement fine-tuning of the generative model and enhancing the emotional continuity of images. Furthermore, we design a \textit{self-promotion textual feedback} framework, in which the LVLM iteratively analyzes the emotional content of generated images and adaptively produces refinement suggestions for the next-round prompt, improving the emotional fidelity with fine-grained content. 
Extensive experimental results demonstrate that our approach effectively generates high-quality images with the desired emotions, outperforming existing state-of-the-art methods on both our custom dataset and public dataset.
\end{abstract}

% keywords can be removed

\section{Introduction}
\begin{figure*}[t]
    \centering
    \includegraphics[width=1\linewidth]{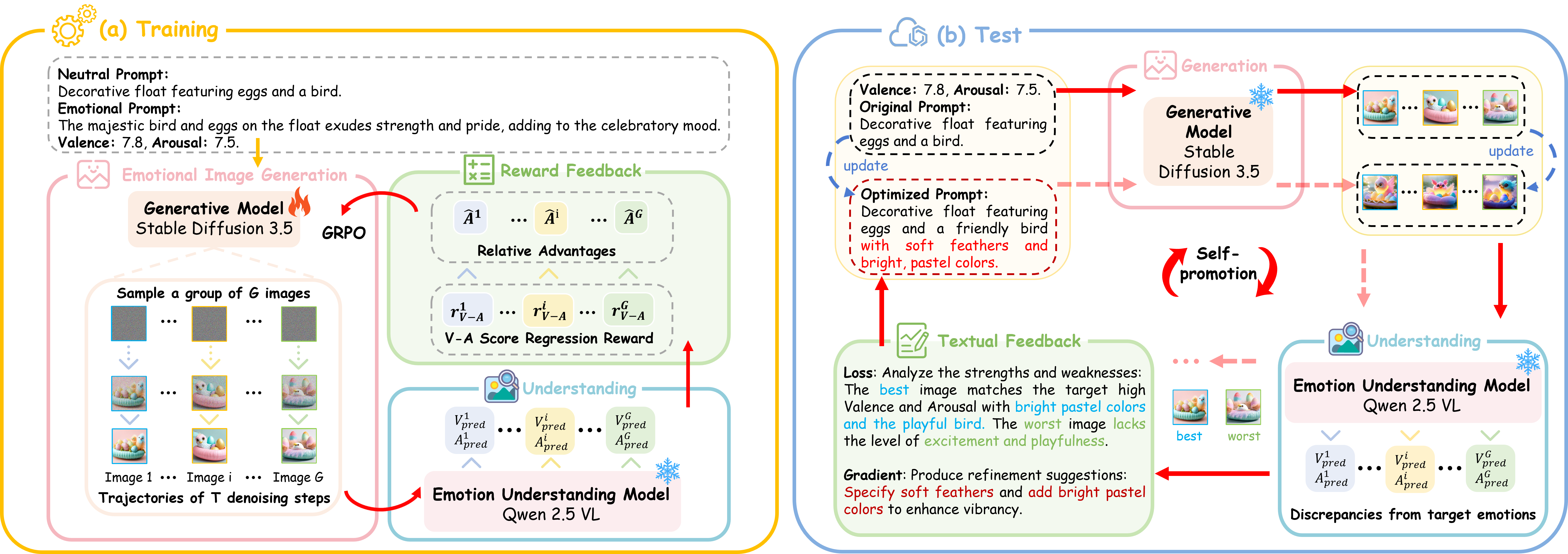}
    \caption{The framework of the EmoFeedback\textsuperscript{2}. During training, given V-A scores, neutral and emotional prompts, the generative model produces a set of images. The emotion understanding model then evaluates the images to provide reward feedback. During testing, the emotional prompt is omitted in C-EICG, and the model iteratively generates textual feedback to refine the prompts.}
    \label{fig:framework}
\end{figure*}
Emotions play a crucial role in shaping our perception and understanding of the world, deeply influencing how we interact with our environment~\citep{chainay2012emotional,yang2018weakly}. Among the many stimuli that evoke emotions, visual cues stand out as particularly powerful due to their intuitiveness and richness of information. Researchers have extensively explored the field of Visual Emotion Analysis (VEA)~\citep{borth2013large, megalakaki2019effects,rao2020learning,wang2022systematic} to investigate the complex interplay between visual content and human emotions. 
In recent years, the rapid advancement of generative models~\citep{ho2020denoising,rombach2022high,esser2024scaling} has enabled them to produce visual content with impressive quality based on textual descriptions. In content creation, incorporating emotional elements is often more effective in engaging and resonating with audiences. However, studies on models capable of generating images reflecting specific emotions remain limited.

Current methods typically construct an emotion encoding network to derive representations from emotional inputs and guide generative models to produce images expressing corresponding emotions. EmoGen~\citep{yang2024emogen} pioneered emotion-driven image generation using discrete tags (e.g., happy, sad) and a mapping network, but its emotional expressiveness was limited by categorical label space. To overcome this, EmotiCrafter~\citep{dang2025emoticrafter} introduced Continuous Emotion Image Content Generation (C-EICG), replacing discrete inputs with continuous Valence and Arousal (V-A) values~\citep{russell1980circumplex} to enable nuanced emotional control. However, current C-EICG methods face several challenges: \textbf{(1) Lack of emotional feedback from generated images}: Their training objective mainly supervise intermediate affective representations upstream of generative model, while the actual emotions expressed in the generated images are not fed back to the model for optimization. As a result, the model fails to capture the subtle variation of emotions in images, constraining its ability to control emotional continuity.
\textbf{(2) Insufficient adaptability for numerical affective inputs}: Existing methods align continuous affective values with pre-generated emotional texts to make numerical inputs interpretable to generative models. This static value–text alignment is independent of the input image and cannot flexibly adjust emotional semantics according to the specific visual content, leading to limited emotional fidelity.

To overcome the two limitations above, we propose EmoFeedback\textsuperscript{2}, a novel generation-understanding-feedback reinforcement paradigm to provide Large Vision Language Model (LVLM)-based reward and textual feedback for C-EICG. Figure \ref{fig:framework} represents the overall framework of our method.
Specifically, we introduce a multi-task reinforcement learning objective to endow the Qwen2.5-VL-7B-Instruct ~\citep{bai2025qwen2} with emotion understanding ability.
Subsequently, we present an emotion-aware reward feedback strategy to better capture the intrinsic relationship between visual content and emotional expression. The LVLM acts as a reward model to measure the discrepancy between predicted and target emotional values, providing emotional feedback to optimize the Stable Diffusion 3.5-Medium (SD3.5-M)~\citep{esser2024scaling} generative model and strengthen control over emotional continuity. In addition, we propose a self-promotion textual feedback optimization framework to adaptively generate the emotional prompts. In each iteration, SD3.5-M produces multiple candidate images, from which the most and least emotion-aligned samples are selected for comparative analysis. Leveraging its chain-of-thought reasoning capability, the LVLM can produce prompt refinement suggestions to enrich emotional descriptions and content details of the next-iteration prompt, thereby improving emotional fidelity and expressiveness.

To summarize, our main contribution can be listed as:

\begin{itemize}
\item We propose a generation-understanding-feedback paradigm for C-EICG, exploiting the reasoning ability of LVLM to provide reward and textual feedback for high-quality and emotionally continuous image generation.
\item We introduce an emotion-aware reward feedback strategy that leverages LVLM to assess the generated images and deliver emotional reward to drive the optimization, enabling continuous and precise emotional control.
\item We design a self-promotion textual feedback framework to analyze the generated content and adaptively optimize the emotional prompts, enhancing emotional fidelity through iterative refinement of content details.
\item We construct a dataset with continuous V-A values, emotion categories, and textual descriptions. Extensive experiments show that our method outperforms existing techniques on both our dataset and public benchmark.
\end{itemize} 

\section{Related Works}
\subsection{Visual Emotion Analysis} 
VEA aims to computationally recognize emotions in images and videos. Early work emphasized discrete categories~\citep{yang2020image, yang2022emotion, xu2022mdan}, but the emerging continuous models highlight dimensions such as arousal, valence, and dominance~\citep{kollias2022abaw, toisoul2021estimation}. Recent studies integrate contextual cues from posture, objects, and scenes~\citep{kosti2017emotion, kragel2019emotion}, achieving strong performance. The central inquiry, what evokes visual emotions, has been explored through low-level (color, texture) and high-level (content, style) features, with contributions such as SentiBank~\citep{borth2013large} and MldrNet~\citep{rao2020learning}. These efforts lay the foundation for generative approaches that embed emotions directly into visual content.
\subsection{Emotional Image Generation}
Most of the previous works in EICG can be grouped into color-based~\citep{chen2020image,liu2018emotional,yang2008automatic,peng2015mixed}, and style-based~\citep{fu2022language, sun2023msnet, weng2023affective}. Recently, EmoGen~\citep{yang2024emogen} pioneered the Emotion Image Content Generation task by generating images based on discrete emotion tags (e.g., happy, sad). The model presents a mapping network to transform abstract emotions into concrete concepts. While groundbreaking, this approach is restricted by the narrow scope of categorical emotion labels, which fail to capture nuanced affective states. To address this limitation, EmotiCrafter~\citep{dang2025emoticrafter} introduced the C-EICG task, along with an emotion-embedding network that injects continuous Valence (V) and Arousal (A) values~\citep{russell1980circumplex} into text prompts to enable smooth, emotion-driven image variation. Meanwhile, EmoEdit~\citep{yang2025emoedit} constructed paired datasets of emotional and original images, designing an Emotion Adapter to mediate interactions between target emotions and input visuals. 
Different from previous works, our method incorporates emotional feedback from outputs to optimize the model and adaptively enrich emotional texts according to the image content.
\section{Method}
\begin{figure*}[t]
    \centering
    \includegraphics[width=1\linewidth]{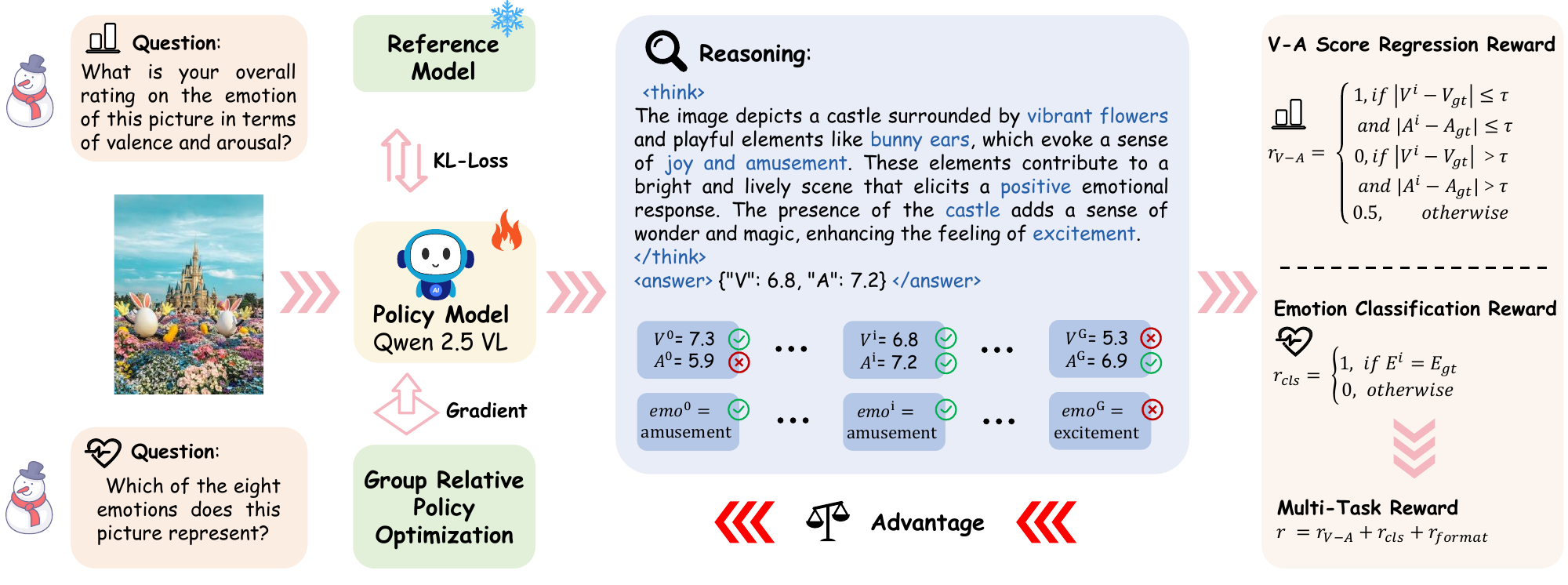}
    \caption{The Emotion Understanding Model Training Process. The training image is input into the emotion understanding model to predict the V-A scores and emotion labels. The set of outputs is then fed into the designed reward functions to calculate the reward. The GRPO algorithm finally derives the advantage and loss to optimize the model.}
    \label{fig:VLM_pipeline}
\end{figure*}
\subsection{Emotion Understanding Model}
LVLMs benefit from large-scale pretraining on diverse image–text pairs that frequently contain affective descriptions, enabling them to learn rich associations between visual patterns and emotional semantics. This provides a strong foundation for emotion understanding. Leveraging the chain-of-thought reasoning capability of LVLM, we introduce a multi-task reinforcement learning objective that jointly optimizes V-A regression and emotion classification. These complementary tasks promote synergistic reasoning to capture both coarse-grained emotional semantics and fine-grained affective intensity. We adopt GRPO~\citep{shao2024deepseekmath} to fine-tune Qwen2.5-VL-7B-Instruct~\citep{bai2025qwen2}. Figure \ref{fig:VLM_pipeline} presents the training pipeline of the emotion understanding model. As shown, the model effectively attends to key emotion-related cues in the reasoning process. To enhance its understanding accuracy, we design three reward functions.

\textbf{Format Reward}: This reward enforces a structured response with reasoning enclosed in \textless think\textgreater\textless/think\textgreater and a JSON-formatted answer in \textless answer\textgreater\textless/answer\textgreater. The reward is 1 if all format requirements are satisfied and 0 otherwise.
    
\textbf{V–A Score Regression Reward}: This reward guides the model to reason about the degree of emotion along valence and arousal dimensions. For each response $o_i$, the predicted values ($V^i$, $A^i$) are compared with the ground-truth values ($V_\text{gt}$, $A_\text{gt}$). A positive reward is assigned when the discrepancy falls within a predefined threshold $\tau$, allowing acceptable deviations without requiring exact matches.
    
\textbf{Emotion Classification Reward}: This task encourages the model to accurately identify the emotion category among eight emotions: amusement, awe, anger, contentment, disgust, fear, excitement, and sadness. We design a binary reward $r_{\text{cls}}$: if the predicted emotion category $E^i$ matches the ground truth label $E_\text{gt}$, the reward is 1; otherwise, it is 0.

\subsection{Emotion-aware Reward Feedback Strategy}
 We design an emotion-aware reward feedback strategy that closes the optimization loop between generated images and target affective inputs. The emotion understanding model (EUM) evaluates the emotions conveyed by each generated image and converts its discrepancy from the target V-A values into a reward feedback for generative model. We optimize SD3.5-M generator using the Flow-GRPO~\citep{liu2025flow} paradigm. For each text prompt, the model performs $T$ denoising steps and generates a group of $G$ images. Each image $\{x_0^i\}_{i=1}^{G}$ is produced through a trajectory $\{(x_T^i, \ldots,x_0^i)\}_{i=1}^{G}$, recording the sequence of latent transitions sampled by the diffusion policy. The EUM assigns the same V-A reward to the final output $x_0^i$, optimizing the transition probabilities at all denoising steps. The group-level rewards are normalized into relative advantages to reflect the intra-group ranking:
\begin{equation}
\hat{A}^{i} = \frac{R\left(x_0^{i}, c\right) - \mathrm{mean}\left(\left\{R\left(x_0^{i}, c\right)\right\}_{i=1}^{G}\right)}{\mathrm{std}\left(\left\{R\left(x_0^{i}, c\right)\right\}_{i=1}^{G}\right)}.
\end{equation}

For each denoising step $t$ along the sampled trajectory, we compute the importance sampling ratio
$r_t^i(\theta)
=
\pi_\theta(x_{t-1}^i \mid x_t^i,c)/
\pi_{\theta_{\mathrm{old}}}(x_{t-1}^i \mid x_t^i,c)$,
where $c$ denotes the text condition. To prevent excessively large policy updates and stabilize training, we define its clipped counterpart as
$\bar r_t^i(\theta)
=
\operatorname{clip}\bigl(r_t^i(\theta),1-\varepsilon,1+\varepsilon\bigr)$.
In addition, a KL divergence penalty weighted by $\beta$ constrains the learned policy to remain close to the reference policy $\pi_{\mathrm{ref}}$. The final optimization objective averages the advantage-weighted policy updates over all sampled trajectories and denoising steps within each group:
{\small
\begin{equation}
\mathcal{J}(\theta)
= \mathbb{E}_{x^{i}\sim\pi_{\theta_{\text{old}}}}
\frac{1}{G}\sum_{i=1}^{G}
  \frac{1}{T}\sum_{t=0}^{T-1}
  \Bigl[\,
  \min\Bigl(
  r_{t}^{i}(\theta)\hat{A}^{i},\;
  \mathrm{clip}\bigl(r_{t}^{i}(\theta),1-\varepsilon,1+\varepsilon\bigr)\hat{A}^{i}
  \Bigr)
  -\beta D_{\mathrm{KL}}\bigl(\pi_{\theta}\parallel\pi_{\mathrm{ref}}\bigr)
  \Bigr],
\end{equation}
}

\subsection{Self-promotion Textual Feedback Framework}
To adaptively optimize emotional prompts according to visual content during inference, we propose a  self-promotion textual feedback framework. At each iteration, multiple candidate images are generated and evaluated, and the best and worst samples are selected according to their discrepancies from the target emotions. Inspired by textual gradient optimization~\citep{yuksekgonul2024textgrad}, our framework performs iterative prompt refinement within the generation-understanding-feedback loop. Instead of updating model parameters with numerical losses and gradients, we formulate the comparative assessment of the best and worst images as a textual “loss”, and regard the directional guidance for visual-content refinement as a textual “gradient”.

The textual feedback optimization consists of three key steps analogous to standard gradient optimization: loss computation, gradient estimation, and variable update. Formally, let $t$ denote the user prompt, $e$ the target emotion, $v$ the generated visual content, $\mathcal{M}$ the LVLM (Qwen2.5-VL-7B-Instruct), and $P$ the prompt function that specifies the instruction for each step. The optimization proceeds as follows:

\textbf{Loss computation}: The LVLM is instructed by $P_{\mathrm{loss}}$ to comparatively analyze the strengths and weaknesses of the best and worst images, producing a textual loss signal that characterizes their differences in emotional expression:
\begin{equation}
\mathcal{L}(e,v) \leftarrow \mathcal{M}\bigl(P_{\mathrm{loss}}(e,v)\bigr).
\end{equation}

\textbf{Gradient estimation}: Guided by $P_{\mathrm{grad}}$, the LVLM reasons over the positive-negative loss to identify the direction of emotion improvement and generate refinement suggestions:
\begin{equation}
\frac{\partial\mathcal{L}}{\partial v}
\leftarrow \mathcal{M}\bigl(P_{\mathrm{grad}}(\mathcal{L}(e,v))\bigr).
\end{equation}

\textbf{Variable update}: Finally, under the instruction $P_{\mathrm{update}}$, the LVLM incorporates the gradient-like suggestions into the user prompt while preserving its original content semantics:
\begin{equation}
t_{\mathrm{opt}} \leftarrow \mathcal{M}\bigl(P_{\mathrm{update}}(\frac{\partial\mathcal{L}}{\partial v},t)\bigr).
\end{equation}

The optimized prompt $t_{\mathrm{opt}}$ is fed into the generative model in the next iteration to produce a new group of images.
By enriching the prompt with additional details and emotional cues, the newly generated images can better align with the desired emotions, while the model parameters remain fixed.

\begin{figure}[t]
    \centering
    \includegraphics[width=1\linewidth]{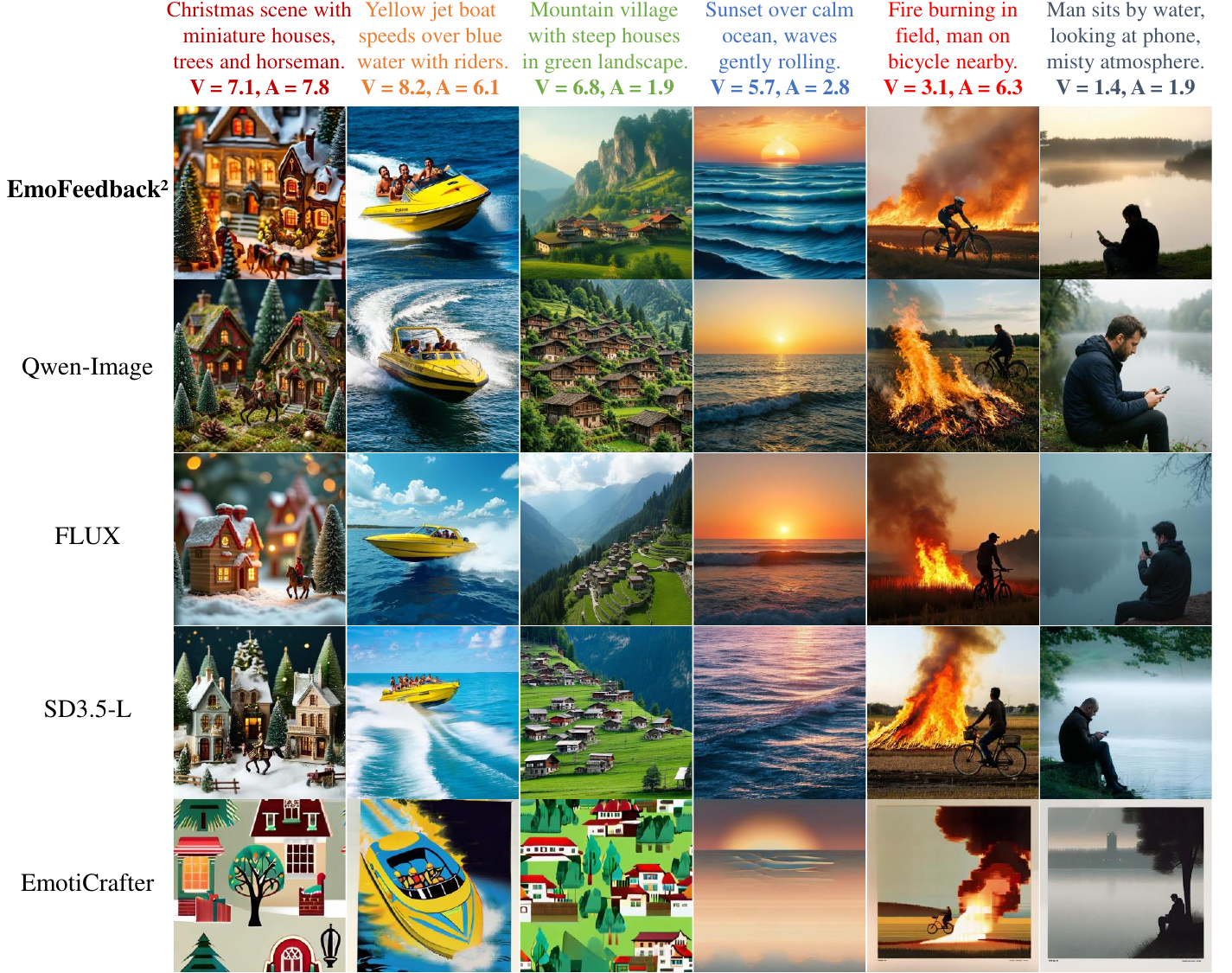}
    \caption{Qualitative comparisons under specific V-A values.}
    \label{fig:Emotion_Image}
\end{figure}

\section{Experiment}
\begin{figure*}[t]
    \centering
    \includegraphics[width=1\linewidth]{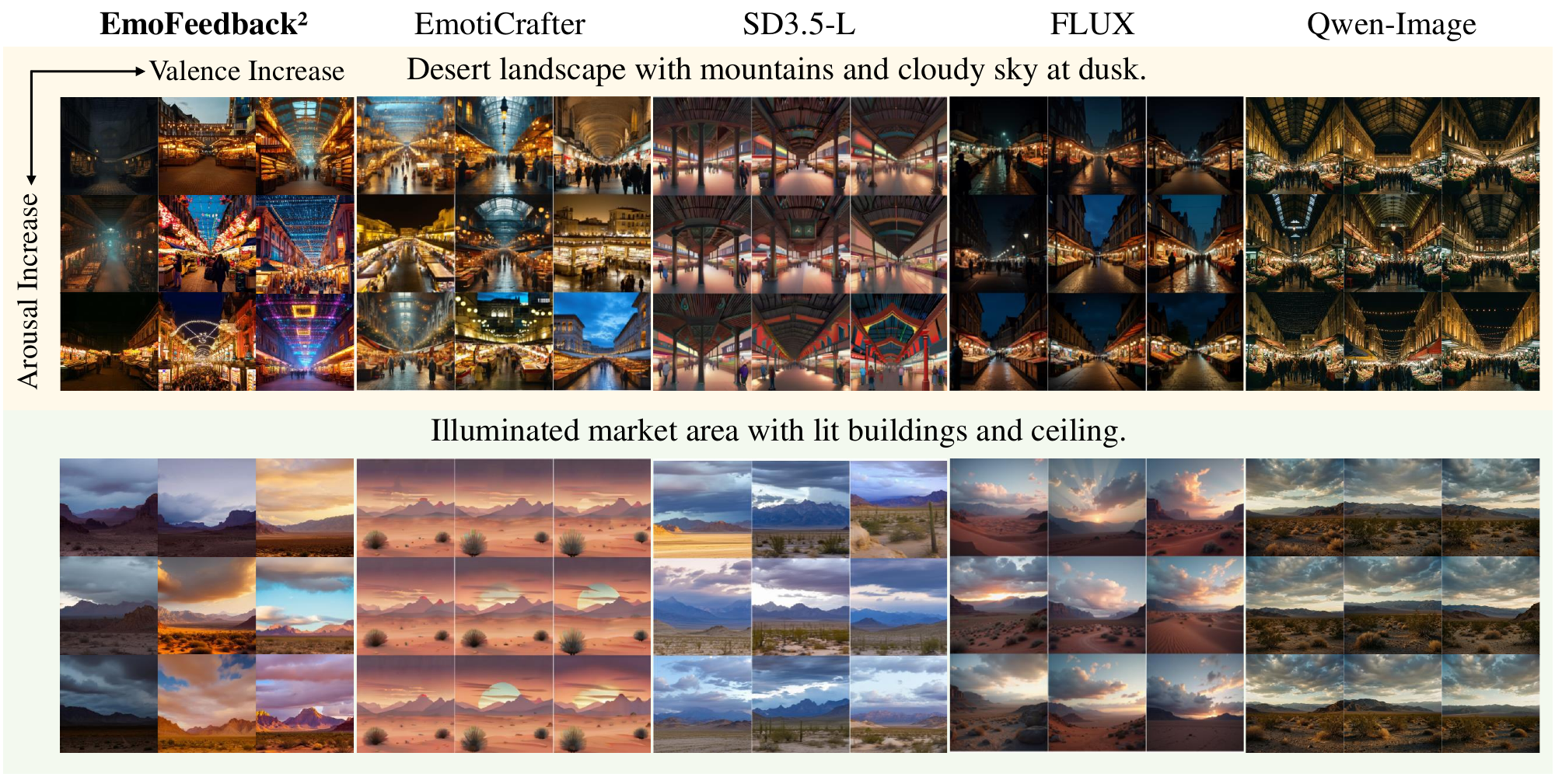}
    \caption{Qualitative comparisons with baselines under varying emotional values.}
    \label{fig:V-A Change}
\end{figure*}
\subsection{Dataset}
\subsubsection{EmoSet-Conti}
We construct a C-EICG dataset annotated with continuous V-A values, emotion categories, and image captions. Built upon EmoSet-118K~\citep{EmoSet}, EmoSet-Conti extends the original dataset with the inputs required for C-EICG. Specifically, the original EmoSet provides only images and eight categorical emotion labels, without textual descriptions or continuous V-A annotations. To address these limitations, we develop a dedicated data-construction pipeline. First, we employ a multimodal large language model to generate neutral captions for each image, forming image-text pairs. For the training set, we additionally generate emotional prompts. Unlike EmotiCrafter, which uses emotional prompts as supervision for emotion-text alignment, our method treats them as inputs to the generative model and optimizes the generator using affective feedback. All textual descriptions are manually reviewed and corrected to ensure their quality. 
We adopt an annotation-verification protocol to construct V-A labels. We derive category-level V-A priors from an affective lexicon~\citep{warriner2013norms} and sample an initial V-A candidate for each image. Each image is then rated by five annotators on a 1-9 scale for Valence and Arousal, without access to its emotion category and initial candidate. The V-A label is obtained by averaging the five ratings. To handle the inherent subjectivity of affective perception, images are further reviewed by three additional annotators if either the rating standard deviation or the label-candidate discrepancy exceeds 1.5. For these samples, the highest and lowest scores are discarded, and the remaining six ratings are averaged to obtain the final V-A label. Finally, we obtain 14,563 training samples and 1,000 test samples.

\subsubsection{EMOTIC}
EMOTIC~\citep{ kosti2019context} is a public emotion dataset with human-annotated V-A labels. We use its complete test split for cross-dataset evaluation without involving any samples in training, validating model generalization to unseen visual domains and independent human annotations.

\subsection{Baseline and Evaluation Metrics}
To effectively evaluate the accuracy of our method in generating emotional images, while also confirming the high quality and aesthetics of the generated images, we select four representative baselines for comparison: \textbf{EmotiCrafter}~\citep{dang2025emoticrafter}, the only existing method specifically designed for C-EICG, and three state-of-the-art text-to-image models from distinct model families, namely \textbf{Qwen-Image}~\citep{wu2025qwenimagetechnicalreport}, \textbf{FLUX}~\citep{flux2024}, and \textbf{Stable Diffusion 3.5 Large} (SD3.5-L)~\citep{esser2024scaling}.
We implement all models on 8 NVIDIA H20 GPUs (96 GB each) under Linux with Python 3.10, PyTorch 2.5.1, and CUDA 12.4. For fair comparison, all methods follow a unified experimental protocol. They use the same
train-test splits and report results over an equal number of generated outputs. Hyperparameters for baselines follow the settings recommended in their original papers. For EmoFeedback\textsuperscript{2}, the V-A reward threshold $\tau$ is set to 0.5, and iterative refinement is performed for three rounds. More details are provided in the Appendix.

We assess our method based on five metrics: V-Error, A-Error, CLIP-Score~\citep{hessel2021clipscore}, CLIP-IQA~\citep{wang2023exploring}, and Aesthetic Score (Aes-Score). \textbf{V-Error} and \textbf{A-Error} evaluate the absolute error of predicted V-A values and the target V-A values, representing the accuracy of emotional expression. Following EmotiCrafter, we adopt the CLIP-based affective evaluator in~\citep{mertens2024findingemo} to predict V-A values. \textbf{CLIP-Score} measures the semantic alignment between a generated image and its text prompt, indicating how faithfully the visual content reflects the intended description. \textbf{CLIP-IQA} evaluates overall perceptual quality, including visual clarity, naturalness, and the absence of noticeable distortions or artifacts. \textbf{Aesthetic Score} estimates subjective visual appeal, reflecting how visually pleasing and artistically attractive the image appears.

\subsection{Results Comparison}
\subsubsection{Qualitative Comparison}
Figure \ref{fig:Emotion_Image} demonstrates the performance of different methods in generating image under specific emotional values. EmoFeedback\textsuperscript{2} excels at preserving prompt content and effectively integrating emotional details. In contrast, EmotiCrafter struggles with conveying emotions in background content and has lower visual quality. SD3.5-L, FLUX and Qwen-Image generate high-quality visuals but fail to accurately depict emotions. For example, under high-arousal conditions, they struggle to convey the intended energy through the subjects’ facial expressions and body poses. Figure \ref{fig:V-A Change} illustrates how EmoFeedback\textsuperscript{2} and baseline methods generate images that evolve with varying V-A values. As shown, the image content generated by SD3.5-L, FLUX and Qwen-Image exhibits little perceptible change with V-A, as these models were not explicitly designed to capture emotional dynamics. EmotiCrafter demonstrates some degree of emotional changes, such as altering texture colors or adding sun elements at high V-A values. However, EmoFeedback\textsuperscript{2} presents much more pronounced emotional expression in terms of background objects, brightness, color tone, and overall atmosphere, making it more effective in achieving emotionally coherent and visually compelling results.

\begin{table*}[t]
\centering
\caption{Performance comparison of different methods. Best results are in \colorbox{pearDark!20}{blue}, and second-best results are in \colorbox{color_green}{green}.}
\label{tab:quantitative_comparison}

\setlength{\tabcolsep}{7pt}
\renewcommand{\arraystretch}{1.15}

\resizebox{1\textwidth}{!}{
\begin{tabular}{llccccc}
\toprule
\textbf{Dataset}
& \textbf{Method}
& \textbf{V-Error $\downarrow$}
& \textbf{A-Error $\downarrow$}
& \textbf{CLIP-Score $\uparrow$}
& \textbf{CLIP-IQA $\uparrow$}
& \textbf{Aes-Score $\uparrow$}
\\
\midrule

\multirow{5}{*}{EmoSet-Conti}
& EmotiCrafter
& 1.179 & 1.485 & 24.011 & 0.753 & 5.235 \\

& SD3.5-L
& 1.032 & 1.271 & 25.209 & 0.834 & 5.335 \\

& FLUX
& 1.141
& 1.310
& \colorbox{color_green}{25.666}
& 0.817
& \colorbox{color_green}{5.569} \\

& Qwen-Image
& \colorbox{color_green}{0.975}
& \colorbox{color_green}{1.128}
& 25.513
& \colorbox{color_green}{0.839}
& 5.494 \\

\rowcolor{gray!20}
& EmoFeedback\textsuperscript{2}
& \colorbox{pearDark!20}{0.510\textsuperscript{*}}
& \colorbox{pearDark!20}{0.767\textsuperscript{*}}
& \colorbox{pearDark!20}{26.873\textsuperscript{*}}
& \colorbox{pearDark!20}{0.865\textsuperscript{*}}
& \colorbox{pearDark!20}{5.588\textsuperscript{*}}
\\

\midrule

\multirow{5}{*}{EMOTIC}
& EmotiCrafter
& 1.056 & 1.335 & 27.072 & 0.909 & 5.430 \\

& SD3.5-L
& 1.204 & 1.040 & 27.704 & 0.930 & 5.302 \\

& FLUX
& 1.027
& \colorbox{color_green}{0.988}
& \colorbox{color_green}{27.877}
& \colorbox{color_green}{0.930}
& \colorbox{color_green}{5.597} \\

& Qwen-Image
& \colorbox{color_green}{0.961}
& 1.009
& 27.563
& 0.928
& 5.546 \\

\rowcolor{gray!20}
& EmoFeedback\textsuperscript{2}
& \colorbox{pearDark!20}{0.802\textsuperscript{*}}
& \colorbox{pearDark!20}{0.683\textsuperscript{*}}
& \colorbox{pearDark!20}{28.136\textsuperscript{*}}
& \colorbox{pearDark!20}{0.938\textsuperscript{*}}
& \colorbox{pearDark!20}{5.623\textsuperscript{*}}
\\

\bottomrule
\end{tabular}
}
\parbox{\textwidth}{
    \scriptsize
    \textit{Note:} \textsuperscript{*} indicates a statistically
    significant improvement over the best-performing baseline
    for each metric under a two-sided paired bootstrap test ($p<0.05$).
}
\end{table*}

\begin{table}[t]
\centering
\caption{
User study on affective alignment and continuity. Best and second-best results are in \colorbox{pearDark!20}{blue} and \colorbox{color_green}{green}.
}
\label{tab:user_study}

\setlength{\tabcolsep}{4.2pt}
\renewcommand{\arraystretch}{1.12}

\resizebox{0.8\textwidth}{!}{
\begin{tabular}{lccccc}
\toprule
\textbf{Metric}
& \textbf{EmotiCrafter}
& \textbf{SD3.5-L}
& \textbf{FLUX}
& \textbf{Qwen-Image}
& \textbf{EmoFeedback$^2$} \\
\midrule

\multicolumn{6}{l}{\textit{(a) Affective Alignment}} \\
Human V-Error $\downarrow$
& 1.02
& 1.04
& 0.98
& \colorbox{color_green}{0.93}
& \colorbox{pearDark!20}{0.62} \\

Human A-Error $\downarrow$
& 1.13
& 0.99
& 0.94
& \colorbox{color_green}{0.90}
& \colorbox{pearDark!20}{0.72} \\

\midrule
\multicolumn{6}{l}{\textit{(b) Affective Continuity}} \\
V-Mono $\uparrow$
& \colorbox{color_green}{71\%}
& 54\%
& 57\%
& 61\%
& \colorbox{pearDark!20}{87\%} \\

A-Mono $\uparrow$
& \colorbox{color_green}{67\%}
& 52\%
& 55\%
& 59\%
& \colorbox{pearDark!20}{84\%} \\

Transition Smoothness $\uparrow$
& 3.46
& 3.52
& 3.55
& \colorbox{color_green}{3.61}
& \colorbox{pearDark!20}{4.20} \\

Content Consistency $\uparrow$
& 3.78
& 4.25
& 4.18
& \colorbox{color_green}{4.27}
& \colorbox{pearDark!20}{4.31} \\

\midrule
\multicolumn{6}{l}{\textit{(c) Generation Quality}} \\
Quality Preference $\uparrow$
& 6\%
& 16\%
& \colorbox{color_green}{25\%}
& 23\%
& \colorbox{pearDark!20}{30\%} \\

\bottomrule
\end{tabular}
}
\end{table}

\begin{table}[t]
\centering
\caption{
Ablation study of the EUM.
Panel (a) examines model capacity, while Panel (b) analyze reward formulation and task supervision. Best results are highlighted in
\colorbox{pearDark!20}{blue}.
}
\label{tab:understanding_ablation}

\setlength{\tabcolsep}{4.0pt}
\renewcommand{\arraystretch}{1.10}

\resizebox{0.7\textwidth}{!}{
\begin{tabular}{lccccc}
\toprule
\textbf{Size}
& \textbf{Reward}
& \textbf{Supervision}
& \textbf{V-Error $\downarrow$}
& \textbf{A-Error $\downarrow$}
& \textbf{Avg. Error $\downarrow$}
\\
\midrule

\multicolumn{5}{l}{
\textit{(a) Model Capacity}
}
\\

3B
& Thres.
& Multi-task
& 0.628
& 1.217
& 0.923
\\

\midrule

\multicolumn{5}{l}{
\textit{(b) Reward Formulation $\times$ Task Supervision}
}
\\

7B
& Conti.
& Regression
& 0.864
& 0.960
& 0.912
\\

7B
& Conti.
& Multi-task
& 0.819
& 0.896
& 0.858
\\

7B
& Thres.
& Regression
& 0.579
& 0.812
& 0.696
\\

\rowcolor{gray!20}
7B
& Thres.
& Multi-task
& \colorbox{pearDark!20}{0.521}
& \colorbox{pearDark!20}{0.710}
& \colorbox{pearDark!20}{0.616}
\\

\bottomrule
\end{tabular}
}
\end{table}

\begin{table}[t]
\centering
\caption{
Human perception agreement of affective evaluators.
}
\label{tab:eum_human_agreement}
\resizebox{0.6\textwidth}{!}{
\begin{tabular}{lccc}
\toprule
Metric
& General LVLM
& Affective Evaluator
& Our EUM \\
\midrule
V-CCC $\uparrow$
& 0.481
& 0.754
& \textbf{0.823} \\

A-CCC $\uparrow$
& 0.426
& 0.697
& \textbf{0.781} \\
\bottomrule
\end{tabular}
}
\end{table}

\subsubsection{Quantitative Comparison}
Table~\ref{tab:quantitative_comparison} presents the performance of different methods in generating emotional images on the EmoSet-Conti and EMOTIC test sets. On EmoSet-Conti, our method achieves the lowest V-Error and A-Error, while obtaining the highest CLIP-Score, CLIP-IQA, and Aes-Score. These results demonstrate both the accuracy of our generated images in conveying emotions and their strong image quality. On EMOTIC, our method also delivers the best cross-domain performance, indicating our generalization advantage over SOTA methods on unseen domains. 
We adopt a paired bootstrap test with 10,000 prompt-level resamples to compare EmoFeedback\textsuperscript{2} with the best baseline for each metric. Holm correction is applied across the ten comparisons and all adjusted $p$-values remain below 0.05, confirming that the observed performance gains are  statistically significant.

\subsection{User Study}
We conduct a blinded user study with 20 participants to investigate two questions: (1) whether the emotions conveyed by generated images
align with human perception, and (2) whether varying the target V-A values under the same prompt produces monotonic and continuous affective changes.

\subsubsection{Affective Alignment}
We randomly sample 100 prompts in the test sets, yielding 500 images from five methods. Without access to the target V-A values, 20 participants rate the perceived Valence and Arousal on a 1-9 scale. \textbf{Human V-Error and A-Error} are computed as the absolute differences between the mean human-perceived scores and the target values. For the five outputs of each prompt, participants select the image with the highest overall
quality. \textbf{Quality Preference} is calculated as the percentage of prompts for which each method receives the most votes. The positions of five images are independently randomized for each participant.

\subsubsection{Affective Continuity}
We further select 50 prompts to evaluate continuous affective control. For each method and prompt, we generate a $3\times3$ affective grid by taking $V,A\in\{2,5,8\}$, while the random seed and all other generation settings are kept fixed. To avoid ordering-induced bias, the individual images are first presented in random order for perceived V or A scoring. The complete grid is then displayed, and participants rate its emotional \textbf{Transition Smoothness} and \textbf{Content Consistency} on a 1-5 scale. Human-perceived V-A monotonicity (\textbf{V-Mono and A-Mono}) are calculated as the proportions of adjacent image pairs whose mean perceived affective scores increase in the
intended direction.

As shown in Table~\ref{tab:user_study}, EmoFeedback$^2$ achieves the lowest human-perceived V--A errors, demonstrating that the emotions conveyed by our generated images are better aligned with human perception. It also obtains the highest monotonicity, smoothness, content consistency, and quality preference, confirming that the emotions evolve continuously while preserving the core semantics and overall quality.

\subsection{Ablation Study}
\subsubsection{Emotion Understanding Model}
Our emotion understanding model (EUM) is built upon Qwen2.5-VL-7B-Instruct and jointly optimizes multi-task objective with a thresholded V-A reward. We systematically analyze its design from three perspectives: model capacity, reward formulation, and task supervision. As shown in Table~\ref{tab:understanding_ablation}, increasing the backbone size from 3B to 7B substantially reduces both V-Error and A-Error, indicating that stronger model capacity benefits fine-grained affective understanding. We further compare continuous (Conti.) and thresholded (Thres.) V-A rewards under both regression-only and multi-task objectives. The thresholded formulation consistently yields lower errors,suggesting that tolerating minor deviations mitigates sensitivity to the inherent ambiguity of affective annotations and focuses optimization on meaningful prediction errors. Finally, multi-task optimization improves V-A estimation under both reward formulations, proving that categorical emotion semantics provide complementary cues for continuous affect regression.

To evaluate the agreement between our EUM and human affective perception, we compare our EUM with the CLIP-based affective evaluator and the general Qwen2.5-VL-7B on the images from test sets. For each evaluator, we compare the predicted V-A values with the human-annotated ratings and report the concordance correlation coefficient (CCC) for Valence and Arousal. As shown in Table~\ref{tab:eum_human_agreement}, our EUM achieves the highest V-CCC and A-CCC. These results demonstrate that emotion-specific training enables the EUM to produce affective judgments that are more consistent with human perception, supporting its reliability as the feedback model.

\subsubsection{Reward and Textual Feedback}
\begin{figure}[t]
    \centering
    \includegraphics[width=1\linewidth]{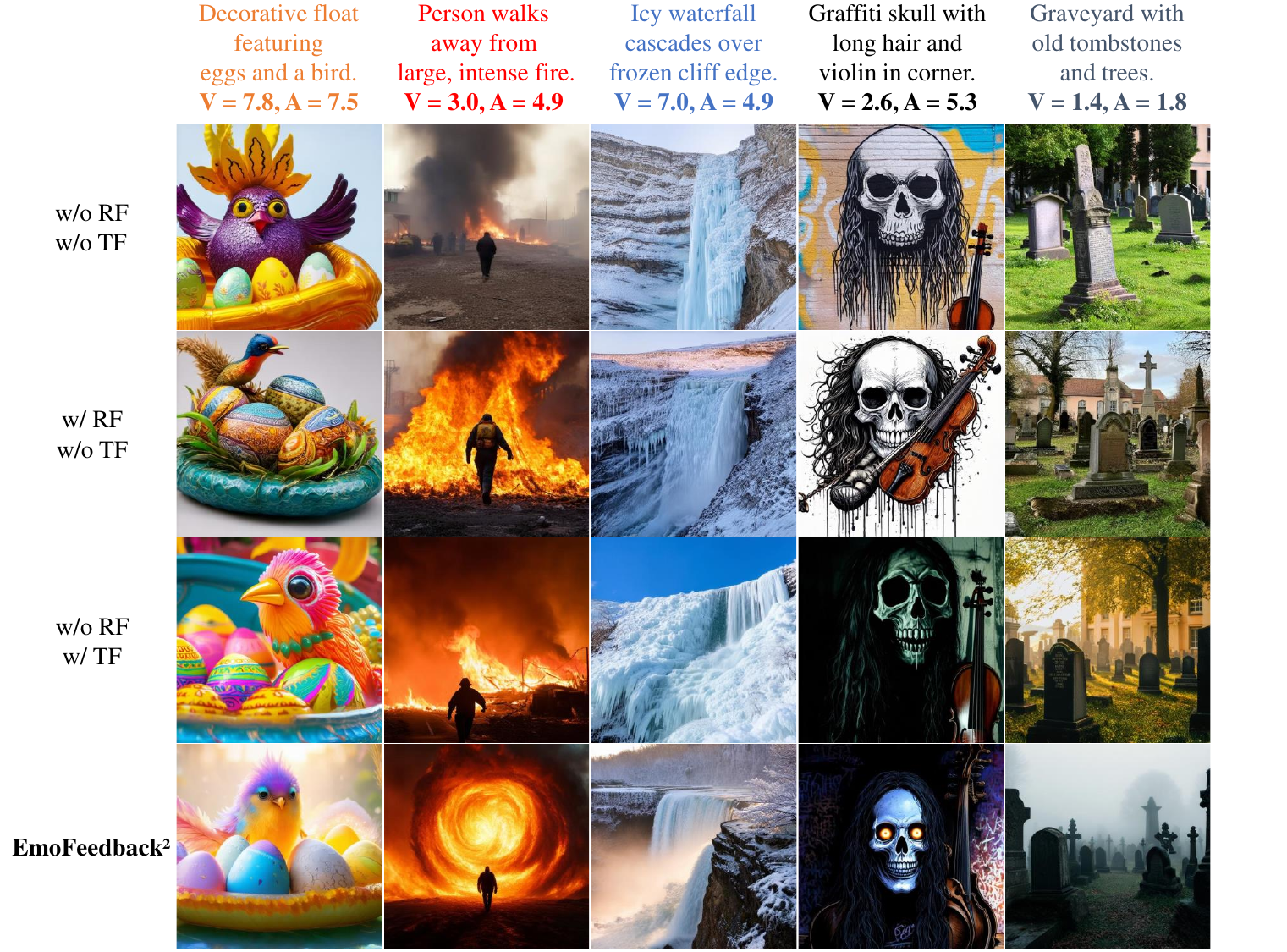}
    \caption{Ablation study on the reward and textual feedback.}
    \label{fig:ablation}
\end{figure}

We evaluate the impact of reward feedback (RF) and textual feedback (TF) in EmoFeedback\textsuperscript{2} on the emotional content of the generated images. As shown in Figure \ref{fig:ablation}, reward feedback enables the image to have emotional content from the initial generation, while textual feedback primarily enriches the details of the generated image to improve emotional expressiveness. 
Table~\ref{tab:feedback_ablation} quantitatively assesses the contributions of RF and TF. Compared with the original SD3.5-M generator, both feedback mechanisms substantially reduce V-Error and A-Error. RF provides larger gains by directly optimizing the generator with image-level affective rewards, whereas TF consistently reduces the errors through content-adaptive prompt refinement. Their combination achieves the best performance, demonstrating the complementary benefits of RF and TF.

\begin{table}[t]
\centering
\caption{
Quantitative ablation of reward feedback (RF) and textual feedback (TF).
Best results are highlighted in \colorbox{pearDark!20}{blue}.
}
\label{tab:feedback_ablation}

\setlength{\tabcolsep}{5pt}
\renewcommand{\arraystretch}{1.10}

\resizebox{0.6\textwidth}{!}{
\begin{tabular}{lccccc}
\toprule
\textbf{Variant}
& \textbf{RF}
& \textbf{TF}
& \textbf{V-Error $\downarrow$}
& \textbf{A-Error $\downarrow$}
& \textbf{Avg. Error $\downarrow$}
\\
\midrule

w/o RF \& TF
& \xmark
& \xmark
& 1.118
& 1.356
& 1.237
\\

TF only
& \xmark
& \cmark
& 0.683
& 0.904
& 0.794
\\

RF only
& \cmark
& \xmark
& 0.622
& 0.861
& 0.742
\\

\rowcolor{gray!20}
Full model
& \cmark
& \cmark
& \colorbox{pearDark!20}{0.510}
& \colorbox{pearDark!20}{0.767}
& \colorbox{pearDark!20}{0.639}
\\

\bottomrule
\end{tabular}
}
\end{table}

\begin{table}[t]
\centering
\caption{
Comparison of different feedback models.
Best results are in \colorbox{pearDark!20}{blue},
and second-best results are in \colorbox{color_green}{green}.
}
\label{tab:feedback_model_analysis}

\setlength{\tabcolsep}{3.5pt}
\renewcommand{\arraystretch}{1.10}

\resizebox{0.7\textwidth}{!}{
\begin{tabular}{lccccc}
\toprule
\textbf{Model}
& \textbf{V-Error $\downarrow$}
& \textbf{A-Error $\downarrow$}
& \textbf{CLIP-Score $\uparrow$}
& \textbf{CLIP-IQA $\uparrow$}
& \textbf{Aes-Score $\uparrow$}
\\
\midrule

\multicolumn{6}{l}{\textit{(a) Preference RM}} \\[-1pt]

PickScore
& 0.617
& 0.849
& 25.515
& 0.790
& 5.337
\\

ImageReward
& 0.635
& 0.872
& 24.933
& 0.723
& 5.326
\\
\midrule
\addlinespace[2pt]
\multicolumn{6}{l}{\textit{(b) General LVLM}} \\[-1pt]

LLaVA-OV
& 0.823
& 0.904
& 26.714
& 0.816
& 5.378
\\
Qwen2.5-VL
& 0.806
& 0.891
& 26.776
& 0.821
& 5.399
\\
\midrule
\addlinespace[2pt]
\multicolumn{6}{l}{\textit{(c) Emotion LVLM}} \\[-1pt]

LLaVA-OV
& \colorbox{color_green}{0.542}
& \colorbox{color_green}{0.778}
& \colorbox{color_green}{26.824}
& \colorbox{color_green}{0.858}
& \colorbox{color_green}{5.427}
\\

\rowcolor{gray!20}
Qwen2.5-VL
& \colorbox{pearDark!20}{0.510}
& \colorbox{pearDark!20}{0.767}
& \colorbox{pearDark!20}{26.873}
& \colorbox{pearDark!20}{0.865}
& \colorbox{pearDark!20}{5.588}
\\

\bottomrule
\end{tabular}
}
\end{table}

\subsubsection{LVLM-based Feedback Paradigm}
To isolate the contribution of emotion-aware LVLM feedback, we compare three families of feedback models: human-preference reward models (RMs), general-purpose LVLMs, and emotion-specialized LVLMs. PickScore~\citep{Kirstain2023PickaPicAO} and ImageReward~\citep{xu2023imagereward} provide human-preference feedback, while Qwen2.5-VL-7B-Instruct and LLaVA-OneVision-1.5-8B-Instruct (LLaVA-OV)~\citep{an2025llavaonevision15fullyopenframework} are used without emotion-specific fine-tuning. We further train LLaVA-OV with the same objective to obtain an emotion-specialized model with a different backbone. All variants follow the same experimental protocol, differing only in the feedback model. As shown in Table~\ref{tab:feedback_model_analysis}, general LVLMs preserve relatively strong image-text alignment and visual quality but yield higher V-Error and A-Error, indicating that general understanding alone is insufficient for precise continuous affective feedback. PickScore and ImageReward achieve lower V-A errors, suggesting that human-preference supervision implicitly captures some affective cues. Nevertheless, without explicit adaptation to continuous V-A estimation, they remain inferior to emotion LVLMs. Finally, the consistent gains across both emotion LVLMs demonstrate that our feedback design is not tied to a particular model architecture.

\subsection{Inference Latency}
Our self-promotion textual feedback framework supports configurable test-time refinement. In each round, the model generates a batch of images, allowing users to either select a satisfactory result immediately or request further refinement for stronger emotional fidelity. Owing to emotion-aware reward fine-tuning, the model already produces competitive results without iterative refinement. In terms of runtime, generating eight images on one H20 GPU takes 11,s, while LVLM-based emotion evaluation and prompt refinement require an additional 9,s. Thus, the minimum latency is 11,s, and each additional refinement round adds 20,s. Generating 8 samples with three iterations require 51,s. Table~\ref{tab:latency} compares the average per-sample inference latency with existing methods on a single H20. Benefiting from the compact 2.5B generator (SD3.5-M), our method maintains competitive efficiency while providing flexible and accurate emotional control, supporting its potential for practical deployment.

\begin{table}[ht]
\centering
\caption{Comparison of inference latency and generator size.}
\label{tab:latency}

\setlength{\tabcolsep}{2.8pt}
\renewcommand{\arraystretch}{1.08}
\resizebox{0.7\textwidth}{!}{
\begin{tabular}{lccccc}
\toprule
\textbf{Metric}
& \textbf{EmotiCrafter}
& \textbf{SD3.5-L}
& \textbf{FLUX}
& \textbf{Qwen-Image}
& \textbf{EmoFeedback\textsuperscript{2}}
\\
\midrule
Latency (s/sample)
& 1.5 & 4.0 & 17.0 & 65.0 & 6.4
\\
Generator Size (B)
& 3.5 & 8.0 & 12.0 & 20.0 & 2.5
\\
\bottomrule
\end{tabular}
}
\end{table}

\section{Conclusion}
In this paper, we introduce EmoFeedback\textsuperscript{2}, an LVLM-based generation-understanding-feedback paradigm for continuous emotional image content generation (C-EICG). Based on the emotion understanding LVLM, we introduce an emotion-aware reward feedback strategy, in which the LVLM calculates emotional feedback from generated images to fine-tune the generative model. At the inference stage, we further propose a self-promotion textual feedback framework to adaptively provide refined emotional prompts for the next-round generation. Extensive experiments prove that EmoFeedback\textsuperscript{2} generates emotionally faithful and high-quality images that vary smoothly with V-A values, outperforming current SOTA methods in both C-EICG and general T2I fields.

\bibliographystyle{unsrtnat}
\bibliography{references}  %%% Uncomment this line and comment out the ``thebibliography'' section below to use the external .bib file (using bibtex) .





% \end{thebibliography}

\clearpage
\appendix
\section*{Appendix}

\section{The Use of Large Language Models}
In this work, we use Large Language Models (LLMs) to aid and polish writing. We utilize LLMs to refine the quality of our manuscript by suggesting more precise terminology. Additionally, we use LLMs to optimize LaTeX templates for figures, tables, and mathematical expressions, significantly reducing the time and effort required for typesetting complex layouts.
\section{Further Details on the Experiment Setup}
\label{app:experiment_setup}

\subsection{Hyperparameters Specification}
In the training of the emotion understanding model, the GRPO generation number $N$ is set to 8, the batch size is set to 16, and we train 5 epochs for convergence. The weight of the KL divergence penalty $\beta$ is set to 1e-3, while the weights $\alpha_1$ and $\alpha_2$ are set to 0.25 and 0.75, respectively. The V-A reward threshold $\tau$ is set to 0.50. We employ AdamW as the optimizer, using an initial learning rate of 1e-6 that linearly decays to 1e-9 during training.
In the training of the generation model, the GRPO generation number $N$ is set to 8. We use a sampling timestep $T$ = 10 and an evaluation timestep $T$ = 25. The image resolution is 512, and the KL ratio is set to 0.1. We set the training process to 1000 steps, and the batch size of every step is set to 16. During the self-promotion textual feedback framework, we set the iteration number of feedback to 3, and generate 8 images every iteration.

\subsection{Self-Promotion Textual Feedback Framework}
\begin{algorithm}[ht]
\caption{Self-Promotion Textual Feedback Framework}
\begin{algorithmic}[1]
\Require
  \Statex original prompt $t_0$, target emotion $e$, generative model $\mathcal{G}(\cdot)$
  \Statex max iterations $I$, prompt functions $P_{\mathrm{loss}}, P_{\mathrm{grad}}, P_{\mathrm{update}}$
\Ensure optimized image $v_{I}$

\State $v_0 \gets \Gen(t_0)$ \Comment{initial image}
\For{$i = 0$ \textbf{to} $I-1$}
    \State $\Loss(e, v_i) \gets \Model\bigl(P_{\mathrm{loss}}(e, v_i)\bigr)$ \Comment{textual loss}
    \State $\Loss_{\text{best}}, \Loss_{\text{worst}} \gets \text{select-best-worst}\bigl(\Loss(e, v_i)\bigr)$
    \State $\Grad \Loss/\Grad v_i \gets \Model\bigl(P_{\mathrm{grad}}(\Loss_{\text{best}}, \Loss_{\text{worst}})\bigr)$ \Comment{textual gradient}
    \State $t_{i+1} \gets \Model\bigl(P_{\mathrm{update}}(\Grad \Loss/\Grad v_i, t_i)\bigr)$ \Comment{prompt update}
    \State $v_{i+1} \gets \Gen(t_{i+1})$ \Comment{new image}
\EndFor
\State \Return $v_{I}$
\end{algorithmic}
\end{algorithm}

\section{LVLM’s Emotional Evaluation Mechanism }

\subsection{Explicit Prompt Guidance}
\label{app:designed_prompts}
As shown in Table \ref{tab:prompt-appendix}, \ref{tab:self-pro-appendix}, the prompts used during training and evaluation explicitly instruct the model to focus on visual elements highly relevant to emotion, such as: “Please consider visual cues such as weather, light, background object, and facial expression in the decision.” The multimodal alignment mechanism of LVLMs enables the model to prioritize these features in visual encoding, which are known to be important in human emotional perception. The explicit prompts can introduce inductive bias that significantly influences the distribution of attention weights in LVLMs. Therefore, by leveraging the model’s strong instruction-following capability, we improve the controllability and interpretability of emotion assessment.
\subsection{Chain-of-Thought Reasoning}
\label{app:chain-of-thought}
Utilizing the reasoning capabilities and hierarchical feature extraction of LVLM to reveal the process of emotional attribution and cue integration. We instruct the model to explicitly output its reasoning steps, making emotion judgment no longer implicitly encoded but expressed through an interpretable reasoning path. Figure 2 provides a concrete example. Before outputting the final emotion scores, the model clearly states its reasoning: (1) identifying key visual elements such as “castle,” “flowers,” and “bunny ears”; (2) interpreting visual attributes such as “bright” referring to color, “playful” referring to style, and “surrounded” referring to composition; (3) linking these elements to emotional implications, such as “amusement,” “positive,” “excitement”. Through hierarchical feature extraction, the LVLM simultaneously captures low-level visual cues and high-level element attributes, and uses cross-modal associations to map visual features into an abstract emotional semantic space. This forms a coherent reasoning process that greatly enhances the reliability and interpretability of our method.

\begin{table}[!htbp]
    \centering
    \renewcommand{\arraystretch}{1.5}
    \caption{Prompts for Different Tasks. The system prompt is shared across all tasks, while task-specific prompts are additionally designed for each task.}
    \label{tab:prompt-appendix}
    \begin{tabular}{p{13.5cm}}
    \hline
     \textbf{System Prompt:} A conversation between User and Assistant. The user asks a question, and the Assistant solves it. The assistant first thinks about the reasoning process in the mind and then provides the user with the answer. The reasoning process and answer are enclosed within <think> </think> and <answer> </answer> tags, respectively, i.e., <think> reasoning process here </think><answer> answer here </answer>. \\
     \hline
     \textbf{Prompt for VA-Value Regression Task:} What is your overall rating on the valence and arousal of this picture? The valence and arousal rating should be a float between 1 and 9, rounded to two decimal places. For valence, 1 represents very sad and 9 represents very happy. For arousal, 1 represents very calm and 9 represents very active. Please consider visual cues such as weather, light, background object, and facial expression in the decision. Return the result in JSON format with the following keys: "valence": The evaluated valence score. and "arousal": The evaluated arousal score. \\
     \hline
     \textbf{Prompt for Emotion Classification Task:} Analyze the given image and decide which of the following eight emotions the image represents: "amusement", "anger", "awe", "contentment", "fear", "disgust", "excitement", and "sadness". Please consider the weather, light, background objects, and facial expression in the decision. Return the result in JSON format with the following keys: "emotion\_class": The detected emotion (or "null" if none). \\
     \hline
    \end{tabular}
\end{table}

\begin{table}[!htbp]
    \centering
    \renewcommand{\arraystretch}{1.5}
    \caption{Prompts for self-promotion textual feedback.}
    \label{tab:self-pro-appendix}
    \begin{tabular}{p{13.5cm}}
    \hline
     \textbf{System Prompt:} You are an expert in image emotion evaluation. You should first think about the reasoning process in your mind and then provide the user with the answer. The emotion metrics to assess images are Valence (V) and Arousal (A): Valence measures how positive or negative the emotion evoked by the image is. A score of 1 indicates extremely negative emotion, while 9 indicates extremely positive emotion. Arousal (A) measures how calming or stimulating the image is. A score of 1 indicates very calm or passive, while 9 indicates very exciting or active. The V and A rating should be two float values between 1 and 9. You will be given a text prompt containing target Valence and Arousal values, along with two images generated from this prompt by a diffusion model, and their corresponding evaluated emotional scores. \\
     \hline
     \textbf{Prompt for self-promotion textual feedback} 1. Analysis: Compare the best image with the worst image. Based on the relationship and differences between their evaluated emotional values and the target values, analyze the strengths and weaknesses of each image. Also, analyze and specify what aspects of the best image need to be modified or improved to make its emotional values closer to the target. Consider aspects such as lighting and brightness, weather and environment, color and composition, characters and objects in your analysis. 2. Optimization: To achieve a new image that aligns more closely with the target emotional values, rewrite and optimize the original text prompt with a more detailed emotional description. The optimized prompt should guide the diffusion model to generate an image that incorporates the modifications and improvements identified in the analysis. The optimized prompt must be richer in content than the original prompt. It should introduce meaningful modifications in aspects such as lighting, brightness, weather, environment, colors, composition, characters, or objects, while preserving the core semantics of the original prompt. 3. Return the answer only as a valid JSON object with exactly two keys: $analysis$: a single string with the comparative analysis, $optimized\_prompt$: the new optimized prompt as a string, in short word format. Do not include any explanations, headings, or Markdown, only return raw JSON. \\
     \hline
    \end{tabular}
\end{table}

\section{Extended Experimental Results}
\subsection{Additional Qualitative Results}

\label{app:extended_experiment}
\begin{figure*}[t]
    \centering
    \includegraphics[width=1\linewidth]{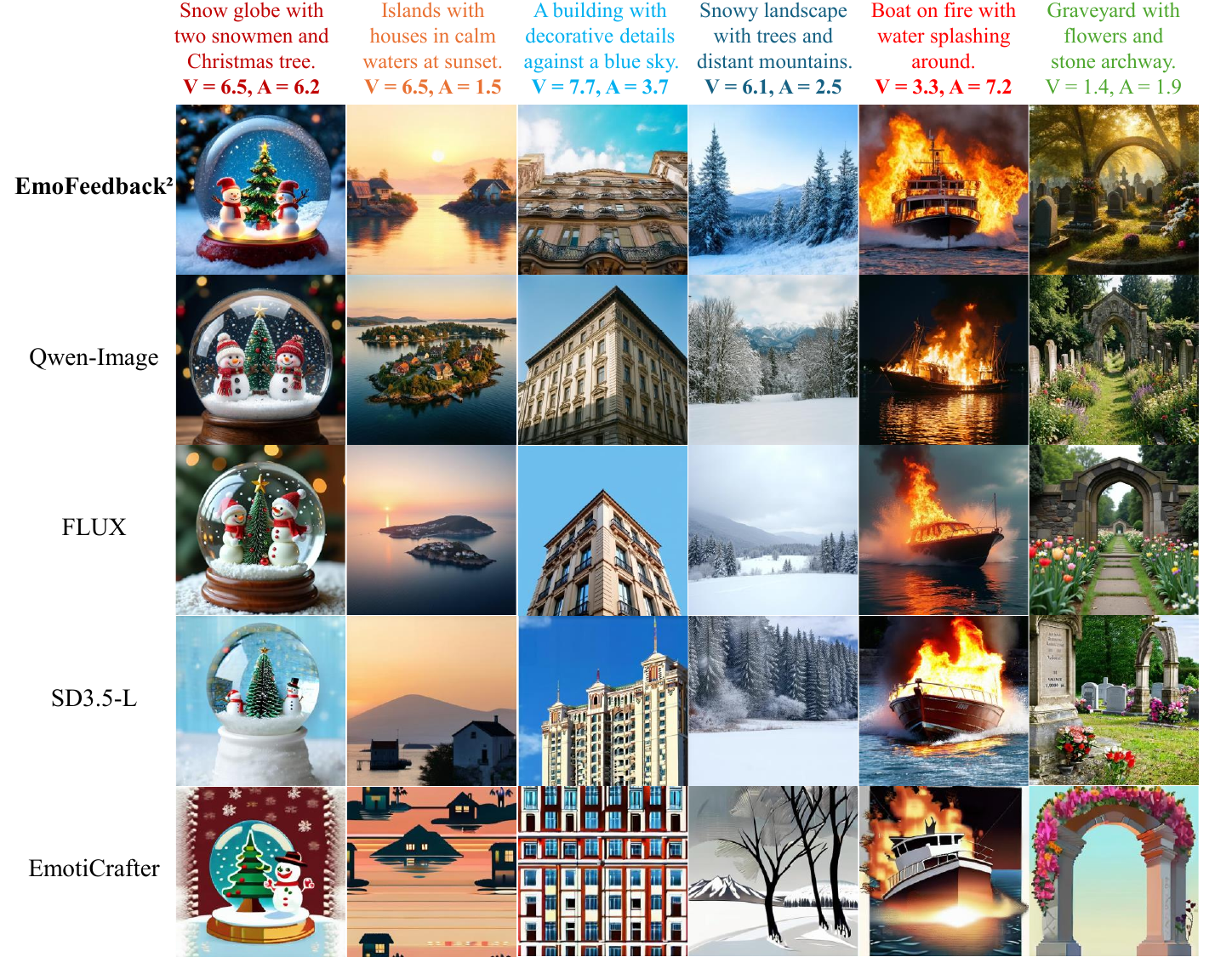}
    \caption{Additional Qualitative results under specific emotional states.}
    \label{fig:Emotion_Image_appedix1}
\end{figure*}

\begin{figure*}
    \centering
    \includegraphics[width=1\linewidth]{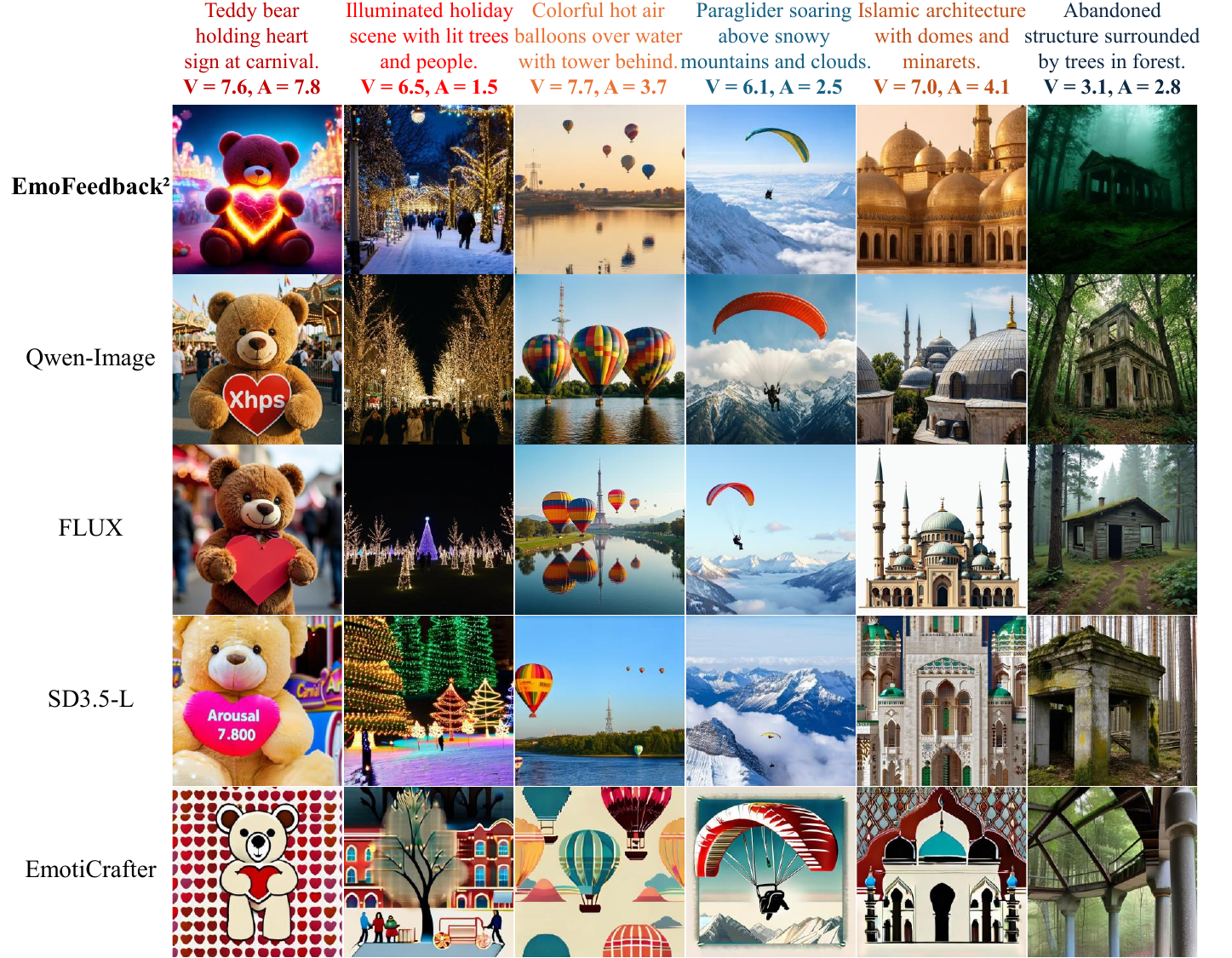}
    \caption{Additional Qualitative results under specific emotional states.}
    \label{fig:Emotion_Image_appedix2}
\end{figure*}

\begin{figure*}
    \centering
    \includegraphics[width=1\linewidth]{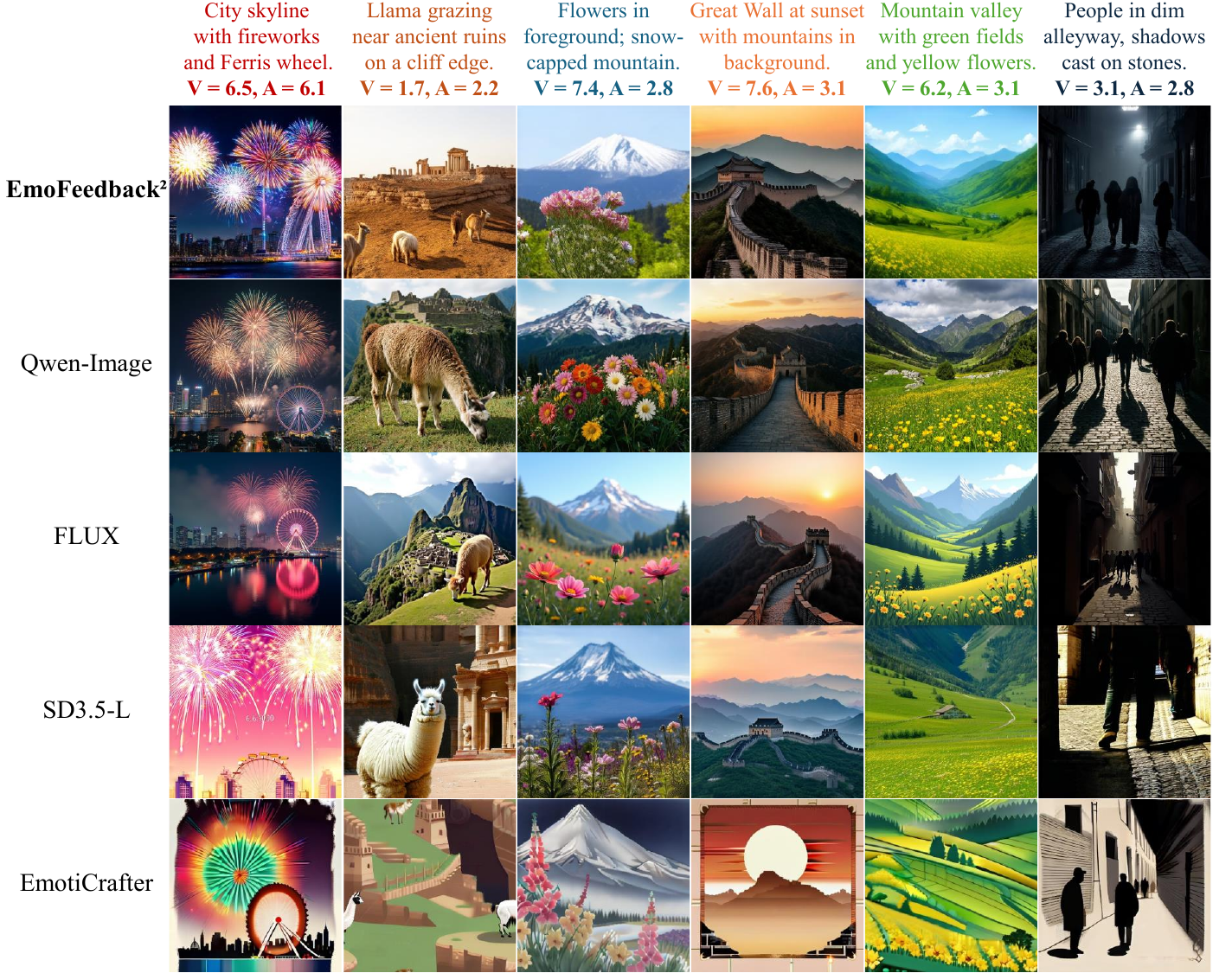}
    \caption{Additional Qualitative results under specific emotional states.}
    \label{fig:Emotion_Image_appedix3}
\end{figure*}

\begin{figure*}
    \centering
    \includegraphics[width=1\linewidth]{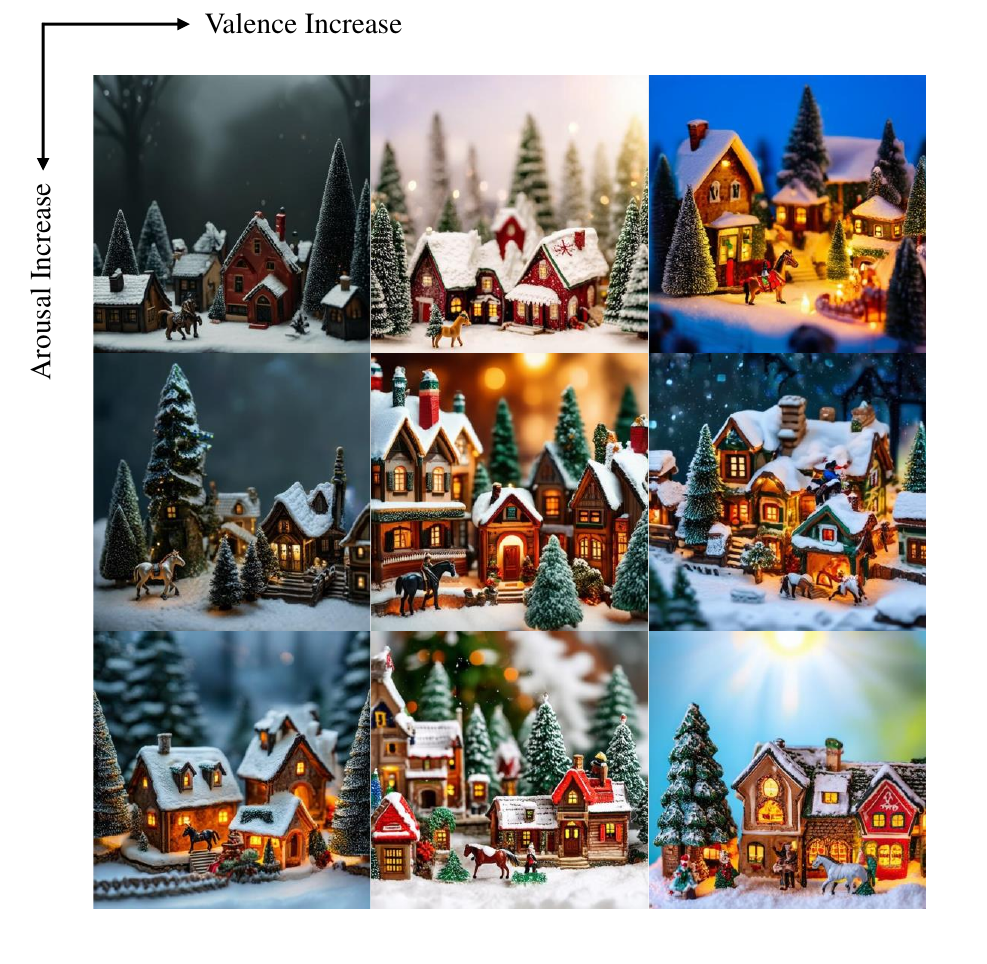}
    \caption{Additional Qualitative results under varying emotional states. The original neutral prompt is "Christmas scene with miniature houses, trees, and a horseman". As valence and arousal increase, the light in this picture turns bright. The light in the cabin makes the atmosphere warmer.}
    \label{fig:Emotion_Image_appedix9_1}
\end{figure*}

\begin{figure*}
    \centering
    \includegraphics[width=1\linewidth]{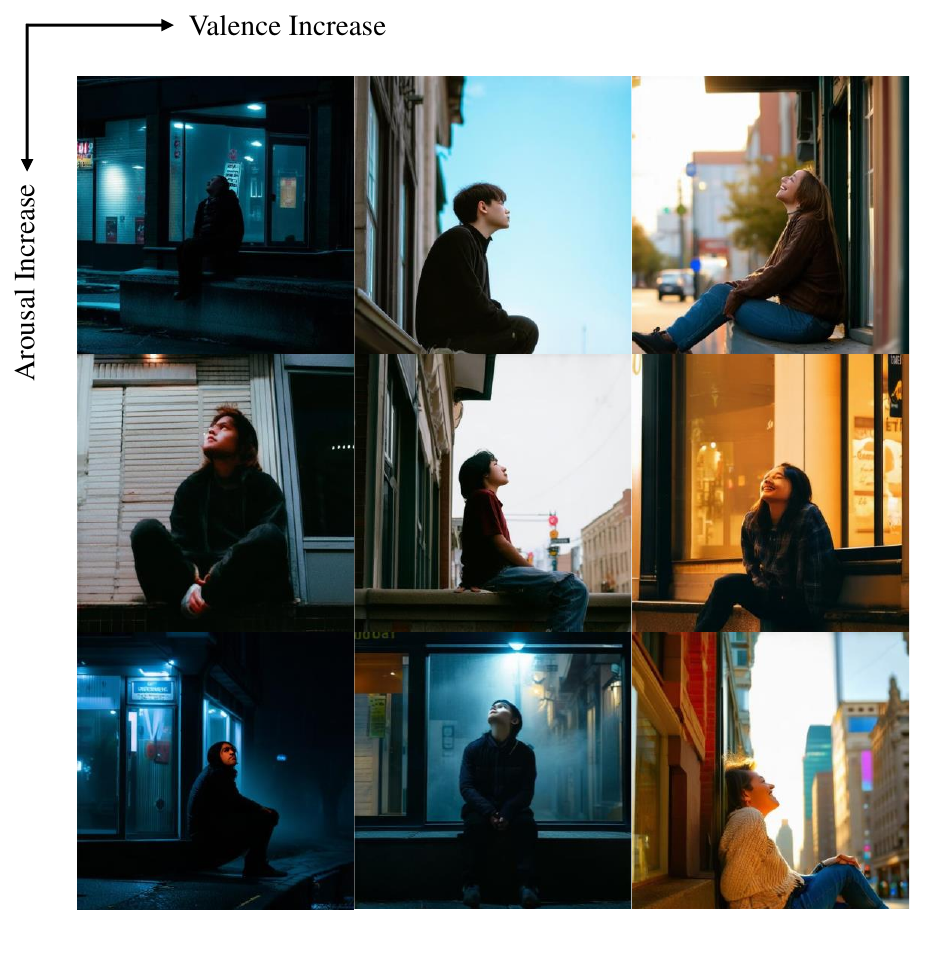}
    \caption{Additional Qualitative results under varying emotional states. The original neutral prompt is "Person sitting on storefront ledge, looking up". As valence and arousal increase, the person's facial expression changes from grave to a laugh. The light in the picture becomes brighter. The city is more colorful.}
    \label{fig:Emotion_Image_appedix9_2}
\end{figure*}

\begin{figure*}
    \centering
    \includegraphics[width=1\linewidth]{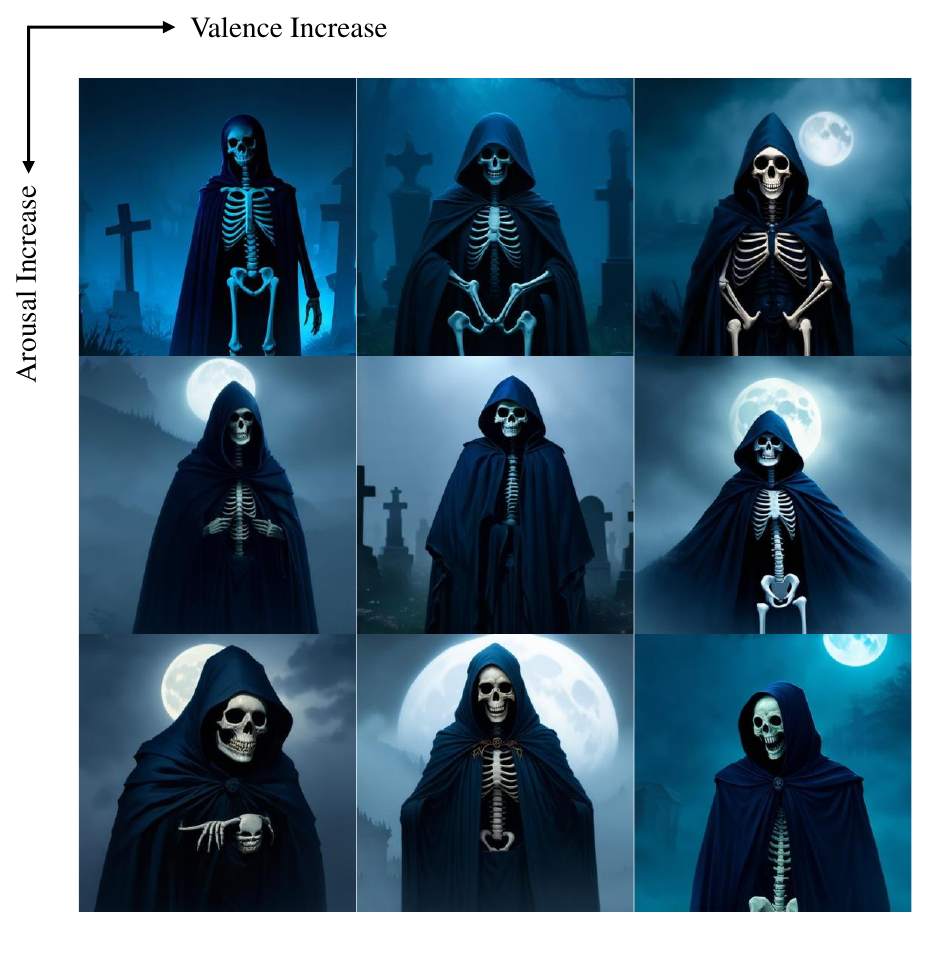}
    \caption{Additional Qualitative results under varying emotional states. The original neutral prompt is "Skull and skeleton wearing a blue hooded cape outside". As valence and arousal increase, the skull's facial expression changed from frightening to amused. The added moon increases the light in the picture.}
    \label{fig:Emotion_Image_appedix9_3}
\end{figure*}

Figures \ref{fig:Emotion_Image_appedix1}, \ref{fig:Emotion_Image_appedix2}, \ref{fig:Emotion_Image_appedix3} qualitatively compare EmoFeedback\textsuperscript{2} and other baselines. Figures \ref{fig:Emotion_Image_appedix9_1}, \ref{fig:Emotion_Image_appedix9_2}, \ref{fig:Emotion_Image_appedix9_3} demonstrate the content variation of the pictures with the change of V and A values in different original emotions.

\subsection{The Textual Feedback}
Table \ref{tab:self_pro_1} and Table \ref{tab:self_pro_2} represent two examples of the analysis process and optimized prompt in our proposed self-promotion
textual feedback framework. 
\begin{table}[!htbp]
    \centering
    \renewcommand{\arraystretch}{1.5}
    \caption{Examples of the self-promotion textual feedback, including analysis and outputs.}
    \label{tab:self_pro_1}
    \begin{tabular}{p{13.5cm}}
    \hline
     \textbf{Original:} Decorative float featuring eggs and a bird-like creature. \\
     \hline
     \textbf{Iteration 1:} Analysis: \textcolor{blue}{The best image has a bright and cheerful color palette, which gives it a positive and lively feeling}. The bird-like creature adds a whimsical element that enhances the overall visual appeal. \textcolor{blue}{The worst image, on the other hand, has a darker and more intense color scheme, making it feel less appealing and more unsettling.} The best image should be slightly more dynamic and vibrant to match the target emotional values.\\
     Optimized prompt: Decorative float featuring eggs and  \textcolor{blue}{ a friendly bird-like creature with soft feathers and bright, pastel colors}. \\
     \hline
     \textbf{Iteration 2:} Analysis: The best image has a higher valence and arousal compared to the worst image. The \textcolor{blue}{bright pastel colors and the playful presence }of the bird-like creature contribute to a more vibrant and stimulating scene. \ \textcolor{blue}{The worst image, despite having a soft and warm feeling, lacks the same level of excitement and playfulness.} \\
     Optimized prompt: Decorative float featuring eggs and a friendly bird-like creature with soft feathers and bright, pastel colors,  \textcolor{blue}{glowing under a soft morning light with a slightly hazy background.} \\
     \hline
    \end{tabular}
\end{table}

\begin{table}[!htbp]
    \centering
    \renewcommand{\arraystretch}{1.5}
    \caption{Examples of the self-promotion textual feedback, including analysis and outputs.}
    \label{tab:self_pro_2}
    \begin{tabular}{p{13.5cm}}
    \hline
     \textbf{Original:} Person walks away from large, intense fire. \\
     \hline
     \textbf{Iteration 1:} Analysis: The best image depicts a person walking away from a large, \ \textcolor{blue}{intense fire, which evokes a strong sense of danger and urgency}. The evaluated emotional values align well with the target emotions. The worst image shows \textcolor{blue}{a person running towards the fire, which adds a dynamic element but may evoke a slightly different emotional response} due to the direction of movement. The best image could benefit from a slight increase in the Arousal value to better match the target emotional values.\\
     Outputs: Person walks away from \textcolor{blue}{a massive, swirling inferno amidst a dark, smoky night sky.} \\
     \hline
     \textbf{Iteration 2:} Analysis: The best image has a slightly lower Arousal value compared to the target, making it more suitable for the given prompt. The worst image has a higher Arousal value, which may not align well with the intended feeling of the prompt. \\
     Outputs: Person walks away from \textcolor{blue}{a massive, swirling inferno amidst a dark, smoky night sky with a glowing, fiery aura enveloping the scene.} \\
     \hline
    \end{tabular}
\end{table}

\section{Broader Impact}
Our paradigm EmoFeedBack\textsuperscript{2}, enables the personalized generation of emotionally evocative images tailored to individual users and holds substantial promise. Leveraging the reasoning and feedback capabilities of Large Vision-Language Models (LVLMs) together with user preference data enables the development of private, user-specific models for emotional image generation. Furthermore, emotional valence-arousal (V-A) features can be decoded directly from electroencephalography (EEG) signals, opening up the possibility for users to enrich images with additional emotional content based on their neural responses to original visual stimuli.

\section{Limitations and Future Work}
Although our method captures continuous affective evolution across varying V-A conditions, it is currently represented through image-level outputs rather than temporally unfolding visual content. Many real-world applications could further benefit from affective changes presented over time. Future work will extend our framework to text-to-video generation conditioned on a target V-A trajectory, aiming to synthesize temporally coherent videos whose visual content gradually develops toward the desired affective states, thereby enabling more immersive affective expression and experiences.

\end{document}